\documentclass[10pt]{article} 

\usepackage{microtype}
\usepackage{graphicx}
\usepackage{subfigure}
\usepackage{booktabs} 

\usepackage{hyperref}
\usepackage{xcolor}
\usepackage{url}
\usepackage{fancyhdr}
\usepackage{standalone}

\usepackage{algorithm}
\usepackage{algcompatible}

\usepackage{amsmath}
\usepackage{amssymb}
\usepackage{mathtools}
\usepackage{amsthm}
\usepackage{bm}
\usepackage{multirow}

\usepackage[preprint]{tmlr}

\usepackage{xcolor}

\newcommand{\trp}{{^\top}} 

\renewcommand{\eqref}[1]{eq.~\ref{eq:#1}}

\newcommand{\figref}[1]{Fig.~\ref{fig:#1}}  
\newcommand{\secref}[1]{Sec.~\ref{sec:#1}}  

\newcommand{\tabref}[1]{Table ~\ref{tab:#1}}  
\newcommand{\algoref}[1]{Algorithm~\ref{algo:#1}}  
 
\newcommand{\propref}[1]{Prop.~\ref{prop:#1}}

\newcommand{\suppsecref}[1]{Sec.~\ref{supp:#1} in Appendix}

\newcommand{\vx}{\mathbf{x}}
\newcommand{\vX}{\mathbf{X}}

\newcommand{\Dat}{\mathcal{D}}

\newcommand{\vk}{\mathbf{k}}

\newcommand{\Em}{\mathbb{E}} 

\newcommand{\LL}{\ensuremath{\mathcal{L}}}

\newcommand{\vy}{\mathbf{y}}

\newtheorem{prop}{Proposition}
\newtheorem{defn}{Definition}[section]

\title{Differentially Private Kernel Inducing Points using features from ScatterNets (DP-KIP-ScatterNet)
for Privacy Preserving Data Distillation}

\author{\name Margarita Vinaroz \email margarita.vinaroz@tuebingen.mpg.de \\
      \addr 
      University of T\"ubingen
      \\
      International Max Planck Research School for Intelligent Systems (IMPRS-IS) \\
      \AND
      \name Mi Jung Park \email mjupa@dtu.dk \\
      \addr Technical University of Denmark
      }

\def\month{MM}  
\def\year{YYYY} 
\def\openreview{\url{https://openreview.net/forum?id=XXXX}} 

\begin{document}

\maketitle

\begin{abstract}
Data distillation aims to generate a small data set that closely mimics the performance of a given learning algorithm on the original data set. The distilled dataset is hence useful to simplify the training process thanks to its small data size. However, distilled data samples are not necessarily privacy-preserving, even if they are generally humanly indiscernible. To address this limitation, we introduce differentially private kernel inducing points (DP-KIP) for privacy-preserving data distillation.
%
Unlike our original intention to simply apply DP-SGD to the framework of KIP, we find that KIP using infinitely-wide convolutional neural tangent kernels (conv-NTKs) performs better compared to KIP using fully-connected NTKs. However, KIP with conv-NTKs, due to its convolutional and pooling operations,  introduces an unbearable computational complexity, requiring hundreds of V100 GPUs in parallel to train, 
 which is impractical and more importantly, such computational resources are inaccessible to many.  
To overcome this issue, we propose an alternative that does not require pre-training (to avoid a privacy loss) and can well capture complex information on images, as those features from conv-NKTs do, while the computational cost is manageable by a single V100 GPU. 
To this end, we propose DP-KIP-ScatterNet, which uses the wavelet features from \textit{Scattering networks} (ScatterNet) instead of those from conv-NTKs, to perform DP-KIP at a reasonable computational cost. 
We implement \textit{DP-KIP-ScatterNet} in -- computationally efficient -- JAX and test on several popular image datasets to show its efficacy and its superior performance compared to state-of-the art methods in image data distillation with differential privacy guarantees.  
\end{abstract}

\section{Introduction}

First introduced by \citet{DD18}, 
data distillation (DD) aims at extracting the knowledge of the entire training dataset to a few synthetic, distilled datapoints. What DD offers is that the models trained on the small number of distilled datapoints achieve high-performance relative to the models trained on the original, large training dataset. DD is a sensible choice for fast, cheaper, and light training of neural network models. Various applications of DD include continual learning \citep{liu2020mnemonics, rosasco2021distilled, sangermano2022sample, wiewel2021condensed, masarczyk2020reducing}, neural architecture search \citep{zhao2021siameseaug, GM, DM}, and more. 

Depending on the similarity metrics chosen for judging how close the small distilled datasets are to the original large datasets, there are different ways to formulate the DD problem. For instance, \citet{GM} formulate it as a gradient matching problem between the gradients of deep neural network weights trained on the original and distilled data. 
 \citet{kip_fc, kip_convnet} formulate it as a kernel ridge regression problem where the distilled data correspond to the kernel inducing points (KIP).  
Regardless of what formulation one takes, the techniques for DD are fast improving, and their application domains are widening. 

Among the many application domains,  \citet{kip_fc} claim that DD is also useful for privacy-preserving dataset creation, by showing distilled images with 90\% of their pixels corrupted while test accuracy with those exhibits  limited degradation. It is true that the distilled images with 90\% of corrupted pixels are not humanly discernible.   
However, their illustration is merely experimental and does not involve any formal definition of privacy.  
Recently, \citet{pmlr-v162-dong22c} attempt to connect DD with differential privacy \citep{Dwork14}, one of the popular privacy notions,  
based on DD's empirical robustness against some known attacks.
%
Unfortunately, the empirical evaluation
of the method and its theoretical analysis contain major flaws, as described in \citet{vitaly22}.

For a provable privacy guarantee,  \citet{DPSGD_GM} apply
\textit{DP-SGD}, the off-the-shelf differential privacy algorithm \citep{abadi_2016}, to optimizing a gradient matching objective 
to estimate a differentially private distilled dataset. More recently, 
\citet{zheng2023differentially} proposes a differentially private distribution matching framework, which further improves the performance of \citet{DPSGD_GM}.

In this paper, we apply DP-SGD to the KIP framework  developed by \citet{kip_fc, kip_convnet} for differentially private data distillation.
%
There are two important reasons we choose to privatize KIP over other existing DD algorithms. 
First, in DP-KIP, the gradients that DP-SGD privatize are the distilled datapoints. Typically, we consider only a few distilled datapoints and therefore \textit{the dimension of the parameters to guard is relatively low}. Consequently, the privacy-accuracy trade-off of DP-KIP is better than that of  the gradient matching framework with DP-SGD \citep{DPSGD_GM}, as the latter needs to  privatize significantly higher dimensional neural network parameters than the former.
 Second, optimizing for distilled data under KIP is \textit{computationally cheaper and converges faster} than that under gradient matching. KIP implements kernel ridge regression (KRR) which involves convex optimization with a closed-form solution and estimating distilled data requires first-order optimization methods,  which is straightforward to apply DP-SGD to and converges relatively fast. However, the gradient matching framework requires computationally expensive bi-level optimization (inner loop updates distilled data and outer loop updates the neural network parameters), which requires a longer training time than KIP. 
In non-DP settings, a longer training time and a higher parameter dimension typically do not matter. However, in DP settings, these incur higher privacy costs, which in turn results in a poorer privacy-utility trade-off. Indeed, for these reasons, we observe that empirically our method outperforms other DP-DD methods at the same privacy level.



Settling with the kernel-based KIP framework, now the question is which features (kernel) to use to produce high quality distilled datasets with privacy guarantees.  
\citet{kip_fc} use the features of infinitely-wide   neural tangent kernel (NTK) from fully connected neural networks. \citet{kip_convnet} significantly improves previous results in \citet{kip_fc} by changing the infinite-width NTK from a fully connected network to a convolutional network. 
 However, kernels formed out of convolutional and pooling layers introduce a significant computational burden, due to the necessity of keeping track of pixel-pixel correlations.
To mitigate this problem, \citet{kip_convnet} use hundreds of V100 GPUs working in parallel to make the optimization computationally feasible.
As a result, using the infinitely-wide ConvNet NTK (ConvNet-NTK) becomes out of reach for researchers who do not have hundred V100 GPUs available at hand. So, we question which features (kernels) to use to produce the results 
better or comparable to the distilled data trained with the features of infinitely-wide ConvNet NTKs, while keeping the computational cost manageable to a single V100 GPU?

The first alternative we consider is an empirical NTK (e-NTK), a finite-dimensional approximation to the infinite-width NTK. e-NTK is known to provide useful approximations of the training dynamics and generalization properties of a given deep neural network \citep{deep_info_prop, xiao2018dynamical, xiao2020disentangling}. It has been observed that as the width of the neural network increases, the e-NTK at initialization approaches the infinite-width NTK. Therefore, we empirically test whether the performance of convolutional e-NTKs is comparable to that obtained by the convolutional infinitely-width NTK. 

The second alternative we consider is  a kernel using the features from a \textit{Scattering Network}, often called ScatterNet \citep{scatternet_2015}. ScatterNet is a feature extractor formed by the combination of predefined complex wavelet filters and non-linear operations followed by a final averaging operator. These handcrafted features exhibit stability to deformations and invariance to translations, making them a suitable choice for image classification tasks such as DD, where the goal is to learn distilled images that minimize the L2-norm between true and predicted labels. Notably,  \citet{tramer2021differentially} previously studied the efficacy of these features in image \textit{classification tasks} with differential privacy guarantees. We attempt to use ScatterNet features for data distillation with differential privacy guarantees. We use an untrained ScatterNet (i.e., at its initialization) to avoid privacy costs for pre-training the network.

The last alternative we consider is a kernel using  \textit{perceptual features} (PFs). PFs are defined as the concatenation of a (pre-trained) deep neural network activations for a given input. They have been widely used in different tasks such as transfer learning \citep{Yosinski_2014, huh2016makes}, super-resolution \citep{Johnson_2016} or image generation \citep{santos_2019, harder2023pretrained}. Although DD task is different from those mentioned beforehand, PFs have been shown to extract useful information from input images and therefore we investigate if the use of these features can contribute to improving the performance of KIP. We use an untrained ResNet-18 (i.e., at its initialization) to avoid privacy costs for pre-training the network.   

We show that KIP with ScatterNet features outperforms e-NTK features and perceptual features, achieving similar performance to the infinite-width convolutional NTK, in non-DP settings. Therefore, we incorporate ScatterNet features in DP-KIP. To separate ours from the original KIP, we call our algorithm by \textit{DP-KIP-ScatterNet}. Our experiments show that DP-KIP-ScatterNet significantly outperforms DP-gradient matching by \citet{DPSGD_GM} and DP-distribution matching by \citet{zheng2023differentially}, when tested on several benchmark image classification datasets. 

Before moving on to describe our method, we summarize our contributions
\begin{itemize}
    \item We propose a DP data distillation framework based on KIP, which uses DP-SGD for privacy guarantees in the resulting distilled data. This itself is a mere application of DP-SGD to an existing data distillation method. However,  motivated by the unbearable computational costs in using the infinite-width convolutional NTKs, we look for alternative features and empirically observe that the features from ScatterNets are the most useful in distilling image datasets, evaluated on classification tasks.
    As a result, DP-KIP-ScatterNet significantly outperforms state-of-the-art DP data distillation methods.  
    \item We further improve the performance of DP-KIP-ScatterNet by considering varying levels of pixel corruption rates. By corrupting some portion of pixels with some noise, we reduce the dimension of distilled data, which results in better privacy-accuracy trade-offs. 
    \item Unlike other most existing data distillation papers that focus on distilling \textit{image data only}, we also test DP-KIP, where we use the original features\footnote{For this task, we use the original inifinite-width fully connected NTKs since tabular data would not benefit from  those
features engineered for image data
such as PFs and ScatterNet features.} in KIP, i.e., the infinitely-wide fully-connected NTKs, on distilling \textit{tabular datasets}. DP-KIP with a relatively small number of distilled samples (only 10 samples/class) significantly outperforms previous DP data generation models at the same privacy level. 
\end{itemize}

\section{Background}

In the following section,  we review the Kernel Inducing Points (KIP) algorithm, different kernel functions we propose to use in KIP (infinite-width NTK, e-NTK, ScatterNet features, PFs) and  differential privacy (DP).

\subsection{KIP}

We start giving some key definitions and describing KIP algorithm from \citet{kip_fc}.
\begin{defn}\label{defn:weakly_close}
 Fix a loss function $l$ and let $f, \tilde{f}  : \mathbb{R}^{D} \rightarrow \mathbb{R}^{C}$ be two functions. Let $\xi \geq 0$. Given a distribution $\mathcal{P}$ on $\mathbb{R}^{D} \times \mathbb{R}^{C}$ , we say $f$ and $\tilde{f}$ are weakly $\xi$-close with respect to
$(l, \mathcal{P})$ if:
\begin{align}\label{eq:weakly_close}
\left| \Em_{(x,y) \in \mathcal{P}} l(f(x),y)- \Em_{(x,y) \in \mathcal{P}} l(\tilde{f}(x),y) \right| \leq \xi.
\end{align}
\end{defn}

\begin{defn}\label{defn:dataset_approx}
 Let $\mathcal{D}$ and $\tilde{\mathcal{D}}$ be two labeled datasets in $\mathbb{R}^{D}$ with label space $\mathbb{R}^{C}$ and $A, \tilde{A}$ be two fixed learning algorithms. Let  $\xi \geq 0$. Given the resulting model $A_{\Dat}$ obtained after training $A$ on $\Dat$,  the resulting model $\tilde{A}_{\tilde{\Dat}}$ obtained after training $\tilde{A}$ on $\tilde{\Dat}$, we say $\tilde{\mathcal{D}}$ is a $\xi$-approximation of $\mathcal{D}$ with respect to $(\tilde{A}, A, l, \mathcal{P} )$ if $\tilde{A}_{\tilde{\Dat}}$ and $A_{\Dat}$ are weakly $\xi$-close with respect to $(l,\mathcal{P})$, where $l$ is a loss function and $\mathcal{P}$ is a distribution on $\mathbb{R}^{D} \times \mathbb{R}^{C}$. 
\end{defn}

In data distillation, the goal is to find a small dataset $\Dat_s$ that is $\xi$-approximation to a large, original dataset $\Dat_t$ drawn from a distribution $\mathcal{P}$ with respect to a learning algorithm $A$ and a loss function $l$:
\begin{align}\label{eq:approx}
\left| \Em_{(x,y) \in \mathcal{P}} l(A_{\Dat_t}(x),y)- \Em_{(x,y) \in \Dat_t}l(A_{\Dat_s}(x),y) \right| \leq \xi.
\end{align}
In KIP, the loss $l$ is a classification accuracy in terms of the L2-distance between true labels and predicted labels; and the learning algorithm $A$ is kernel ridge regression (KRR).  

Consider a target dataset $\mathcal{D}_{t} =  \{ (\vx_{t_{i}}, y_{t_{i}}) \}_{i=1}^{n}$ with input features $\vx_{t_{i}} \in  \mathbb{R}^{D}$ and scalar labels $y_{t_{i}}$. Given a kernel $k$ 
, the KIP algorithm  constructs a small distilled dataset $\mathcal{D}_{s} =  \{ (\vx_{s_{j}}, y_{s_{j}}) \}_{j=1}^{m}$, where $\vx_{s_{j}} \in  \mathbb{R}^{D}$, and scalar labels $y_{s_{j}}$ and importantly $m \ll n$,  such that its performance on a classification task approximates the performance of the target dataset for the same task. Note that \citet{kip_fc} call $\Dat_s$ ``support" dataset (hence the subscript ``s"). In this paper, we will use ``support" and ``distilled" datasets, interchangeably.  

The KIP algorithm, we start by randomly initializing the support dataset and then iteratively refine $\mathcal{D}_{s} $ by minimizing the Kernel Ridge Regression (KRR) loss:
\begin{align}
\label{eq:krr_loss}
\LL(\Dat_s) = \sum_{i=1}^n (y_{t_i} - \vk_{{t_i}s} \trp (K_{ss} + \lambda I  )^{-1} \vy_s )^{2},
\end{align}
with respect to the support dataset $\mathcal{D}_{s}$ 
. Here $\lambda > 0$ is a regularization parameter, $K_{ss}$ is a kernel matrix, where the ($i$, $j$)-th entry is $k(\vx_{s_i}, \vx_{s_j})$, $\vk_{t_is}$ is a column vector where the $j$-th entry is $k(\vx_{t_i}, \vx_{s_j})$, and $\vy_s$ is a column vector where the $j$-th entry is $y_{s_j}$.
During the training phase, the support dataset is updated using a gradient-based optimization method, e.g., using Stochastic Gradient Descent (SGD), until some convergence criterion is satisfied.

   


\subsection{Infinite-width NTK}

Initially, NTKs were proposed to help understand neural networks' training dynamics in a function space. In particular, in the infinite-width limit, the parameters of a neural network do not change from the random initialization over the course of the training and the gradients of the network parameters converge to an infinite-dimensional feature map of the NTK \cite{jacot2018, Lee2019, arora2019, Lee2020}. Characterizing this neural tangent kernel is essential to analyzing the convergence of training and generalization of the neural network.

Based on this finding, \citet{kip_fc} motivate the use of NTK in KIP in the following sense: (a) the kernel ridge regression with an NTK approximates the training of the corresponding infinitely-wide neural network; and (b) it is \textit{likely} that the use of NTK yields approximating $\Dat_t$  by $\Dat_s$  (in the sense of $\xi$-approximations given in \eqref{approx}) for learning algorithms given by a broad class of neural networks. While there is no mathematical proof on point (b), \citet{kip_fc} empirically backed up point (b). 


\subsection{Empirical NTK (e-NTK)}

The empirical NTK (e-NTK) is a finite-dimensional approximation of the infinite-width NTK that provides a way to approximate the dynamics of deep neural networks during training. The e-NTK for a given (finite wide) neural network $f$, with random initial parameters $\theta$ is defined as:

\begin{equation}
    \mbox{e-NTK}(\vx, \vx') = \left[ \frac{  \partial f(\theta, \vx)}{\partial \theta} \right] \left[ \frac{  \partial  f(\theta, \vx')}{\partial \theta} \right]^{\top}
\end{equation}

where  $ \partial f(\theta, \cdot) / \partial \theta$ denotes the neural network Jacobian. 
In our experiments, we considered the e-NTK obtained by using a Lenet \citep{Lenet_1998} and ResNet18 \citep{he_2016_resnet} architectures for grayscale and RGB images respectively.

\subsection{ScatterNet features}

We consider the kernel defined by the inner product of the Scattering Network (ScatterNet) features presented by \citet{scatternet_2015}. ScatterNet is a Scale Invariant Feature Transform (SIFT) feature extractor based on cascades of wavelet transform convolutions  followed by non-linear modulus and a final averaging operator. For a given input image $\vx$, its ScatterNet feature representation is defined as the output of the Scattering Network of depth $J$: 

\begin{equation}
    \phi_{S}(\vx) := A | W_{2}| W_{1} x ||
\end{equation}

where $ W_{2}$ and $ W_{1}$ are wavelet transforms that capture local and global information of the input image, $|\cdot|$ is the modulus operator which computes the magnitude of each wavelet coefficient and guarantees translation-invariant representations, and $A$ is the $2^{J}$ patch-averaging operator that contributes in capturing higher-level patterns and relationships between different parts of the image. The ScatterNet can be seen as a variant of a Convolutional Neural Network (CNN) where the architecture parameters and filters are not learned during the training phase, but are predefined fixed wavelets transforms. 

Following \citet{scatternet_2015}, we use a ScatterNet of depth $J=2$ with wavelets rotated along eight angles. The extracted ScatterNet features for a given image of size $H \times W$ has dimension $(K, \frac{H}{2^{J}},  \frac{W}{2^{J}})$ where $K$ is set to 81 and 243 for grayscale and RBG images respectively.

\subsection{Perceptual features (PFs)}

Here we propose the kernel defined by the inner product of the features extracted from deep convolutional networks (DCNNs), also known as perceptual features (PFs). These features are defined as the concatenation of each layer's output from a fixed deep convolutional neural network. 


In this work we considered the features extracted form a randomly initialized ResNet18 \citep{he_2016_resnet}.

\subsection{Differential privacy (DP)}

Differential privacy is a gold standard privacy notion in machine learning and statistics. Its popularity is due to the mathematical probability. The definition of DP (Definition 2.4 in \cite{Dwork14}) is given below. 

\begin{defn}\label{defn:dp_def}
A randomized mechanism $\mathcal{M}$ is $(\epsilon, \delta)$-differentially private if for all neighboring datasets $\mathcal{D}, \mathcal{D'}$ differing in an only single entry 
and all sets $S \subset range(\mathcal{M})$, the following inequality holds:
$$\mbox{Pr}[ \mathcal{M}(\mathcal{D}) \in S ] \leq e^{\epsilon} \cdot \mbox{Pr}[\mathcal{M}(\mathcal{D'}) \in S] + \delta$$
\end{defn}

The definition states that for all datasets differing in an only single entry, the amount of information revealed by a randomized algorithm about any individual’s participation is bounded by $\epsilon$ and $\delta$ (which is preferably smaller than $1/ |\mathcal{D}|$). In our work, we use the inclusion/exclusion definition for neighbouring datasets and privacy per item or per (training) example as protection granularity.

A common paradigm for constructing differentially private algorithms is to add calibrated noise to an algorithm's output. In our algorithm, we use the \textit{Gaussian mechanism} to ensure that the distilled dataset satisfies the DP guarantee. For a deterministic function $h: \mathcal{D} \rightarrow \mathbb{R}^d$, the Gaussian mechanism is defined by $\tilde{h} (\mathcal{D}) = h(\mathcal{D}) + \mathcal{N}(0,\Delta_{h}^2 \sigma^2 I_{d})$. Here, the noise scale depends on the \textit{global sensitivity} \citep{dwork2006our}  of the function $h$, $\Delta_{h}$, and it is defined as the maximum difference in terms of $l_2$-norm, $\|  h(\mathcal{D}) -  h(\mathcal{D'})\|_2$, for $\mathcal{D}$ and $\mathcal{D'}$ differing in an only single entry and $\sigma$ is a function of the privacy level parameters, $\epsilon$, $\delta$.

Differential privacy has two fundamental properties that are useful for applications like ours: composability  \citep{dwork2006our} and post-processing invariance  \citep{Dwork2006}. Composability ensures that if all components of a mechanism are differentially private, then its composition is also differentially private with some privacy guarantee degradation due to the repeated use of the training data. 
In our algorithm, we use the subsampled RDP composition by \citep{wang2019subsampled}, as they yield to tight bounds for the RDP parameter, $\epsilon_{RDP}$,  for a subsampled mechanism. To switch from the definition of RDP to $(\epsilon, \delta)$-DP, we follow \citep{wang2019subsampled}, which keeps track of the parameters of RDP and converts them to the definition of $(\epsilon, \delta)$-DP through the use of numerical methods.
Furthermore, the post-processing invariance property states that any application of arbitrary data-independent transformations to an $(\epsilon, \delta)$-DP algorithm is also $(\epsilon, \delta)$-DP. In our context, this means that no information other than the allowed by the privacy level $\epsilon, \delta$, can be inferred about the training data from the privatized mechanism.

\section{Our algorithm: DP-KIP and DP-KIP-ScatterNet}

In this section, we introduce our proposed algorithm \textit{differentially private kernel inducing points (DP-KIP)}. The algorithm produces differentially private distilled samples by clipping and adding calibrated noise to the distilled data's gradients during training. \textit{DP-KIP-ScatterNet} is a particular algorithm of DP-KIP which uses the features from Scattering Networks. 

\subsection{Outline of DP-KIP}

Our algorithm is shown in \algoref{alg_dpkip}. We first initialize the distilled (support) dataset $\mathcal{D}_{s}$, where the learnable parameters $\vX_{s} = \{\vx_{s_{j}}\}_{j=1}^{m}$ are drawn from a standard Gaussian distribution, (i.e. $\vx_{s_{j}} \sim \mathcal{N}(0,I_{D})$ for $ \vx_{s_{j}} \in \mathbb{R}^{D}$). 
We generate labels $\vy_s$ by drawing them from a uniform distribution over the number of classes; and fix them during the training for $\vX_{s}$. Note that the original KIP algorithm has an option for optimizing the labels through \textit{Label Solve} given optimized distilled images. 
However, we choose not to optimize for the labels to reduce the privacy loss incurring during training.

At each iteration of the algorithm, we randomly subsample $B$ samples from the target dataset $\mathcal{D}_{t}$, $\mathcal{D}_{t_{B}}$. Given a kernel $k$, we compute the loss given in \eqref{krr_loss}.  
Then, we compute the per target sample gradients with respect to the support dataset, given in 
\begin{align}
\label{eq:grad_krr_loss_persample}
    g(\vx_{t,l}, y_{t,l})&:=\nabla_{\Dat_s} \LL(\Dat_s) \nonumber \\
    &=  \nabla_{\Dat_s} \left(\vy_{t_{l}} - \vk_{t_{l}s} (K_{ss} + \lambda I  )^{-1} \vy_s \right)^{2}.
\end{align}
As in DP-SGD, we ensure each datapoint's gradient norm is bounded by explicitly normalizing the gradients if its $l2$-norm exceeds $C$. In the last steps of the algorithm, the clipped gradients are perturbed by the Gaussian mechanism and averaged as in DP-SGD algorithm and finally, the support dataset is updated by using some gradient-based method (e.g. SGD, Adam).

\begin{algorithm}[t]
   \caption{DP-KIP}
   \label{algo:alg_dpkip}
\begin{algorithmic}
   \STATE {\bfseries Input:} Dataset $\mathcal{D}_{t} = \{ (\vx_{t_{i}}, y_{t_{i}}) \}_{i=1}^{n}$, number of distilled samples to generate $m$, number of iterations $P$, mini-batch size $B$, clipping norm $C$, privacy level $(\epsilon, \delta)$
   \STATE \textbf{Step 1}. Initialize distilled dataset $\mathcal{D}_s =\{ (\vx_{s_{j}}, y_{s_{j}}) \}_{j=1}^{m}$ with $\vx_{s_{i}} \sim \mathcal{N}(0,I_{D})$
   \STATE \textbf{Step 2}. Given a desired level of $(\epsilon, \delta)$-DP, we compute the privacy parameter $\sigma$ using the auto-dp package by \citet{wang2019subsampled}.
   
   \FOR{$p=1$ {\bfseries to} $P$}
   \STATE  \textbf{Step 3}. Randomly subsample $\mathcal{D}_{t_{B}} = \{ (\vX_{t_{B}},  \vy_{t_{B}})\} $
   \STATE \textbf{Step 4}. Compute KRR loss given in \eqref{krr_loss}.
   \STATE  \textbf{Step 5}. Compute per-sample gradients in \eqref{grad_krr_loss_persample}
   for each $l \in t_{B}$.
   
   \STATE  \textbf{Step 6}. Clip the gradients via
   $ \hat{g}(\vx_{l}) = g(\vx_{l}) / \max(1, \| g(\vx_{l}) \|_2 / C) $
   \STATE  \textbf{Step 7}. Add noise: $ \tilde{g} =  \sum_{l=1}^{B} \hat{g}(\vx_{l}) + \mathcal{N}(0,\sigma^2 C^2 I)$.
   \STATE  \textbf{Step 8}. Update distilled dataset $\mathcal{D}_s$ with SGD.
   \ENDFOR
   \STATE {\bfseries Return:} Learned private support dataset $\mathcal{D}_s$

\end{algorithmic}
\end{algorithm}

\propref{dpkip_is_dp} states that the proposed algorithm is differentially private.

\begin{prop}\label{prop:dpkip_is_dp}
The DP-KIP algorithm produces a ($\epsilon, \delta$)-DP distilled dataset.
\begin{proof}
Due to the Gaussian mechanism, the noisy-clipped gradients per sample are DP. By the post-processing invariance property of DP, the average of the noisy-clipped gradients is also DP. Finally, updating the support dataset with the aggregated-noisy gradients and composing through iterations with the subsampled RDP composition \citep{wang2019subsampled} produces $(\epsilon, \delta)$-DP distilled dataset. The exact relationship between  $(\epsilon,  \delta)$, $T$ (number of iterations DP-KIP runs), $B$ (mini-batch size),  $N$ (number of datapoints in the target dataset), and $\sigma$ (the privacy parameter) follows the analysis of \citet{wang2019subsampled}.
\end{proof}

\end{prop}

\subsection{Few thoughts on the algorithm}

\textbf{Support dataset initialization:} The first step in DP-KIP initializes each support datapoint in $\mathcal{D}_s$ to be drawn from the standard Gaussian distribution, $\mathcal{N}(0,I_{D})$. This random initialization ensures that no sensitive information is inherit by the algorithm at the beginning of training. Nevertheless, one can choose a different type of initialization such as randomly selecting images from the training set as in \citep{kip_fc, kip_convnet} and then, privatize those to ensure that no privacy violation incurs during the training process. The downside of this approach is that the additional private step in initialization itself is challenging, since computing the sensitivity for neighboring datasets has no trivial bound on the target dataset and incurs in an extra privacy cost.

\textbf{Clipping effect of the gradients:} In our algorithm, we follow the approach from \citep{abadi_2016} and clip the gradients to have $l_2$-norm $C$. This clipping norm is treated as an hyperparameter since gradient values domain is unbounded a priori. Setting $C$ to a relatively small value is beneficial during training as the total noise variance is scaled down by the $C$ factor. However, the small value may result in a large amount of the gradients being clipped and thus, drastically discard useful or important information. In contrast, setting $C$ to a large value helps preserving more information encoded in the gradients but it yields to a higher noise variance being added to the gradients, which may worsen the learned results. Consequently, finding a suitable $C$ is crucial to maintain a good privacy-accuracy trade-off on DP-KIP algorithm. 
See \suppsecref{clipping_effect} for experimental results of the clipping effect.

\textbf{Regularization parameter:} Following \citet{kip_fc} we set the regularization parameter $\lambda$ in \eqref{krr_loss} to $\frac{1}{m} \lambda \cdot \mbox{tr}(K_{ss})$ where $m$ is the number of datapoints in the distilled dataset $X_{s}$. In this fashion, the regularization parameter is scale-invariant to the kernel function and ensures a similar performance of the algorithm for a wide range of $\lambda$ values.

\textbf{Why not privatize the feature maps?} Since in this work we are considering kernel functions $k$ with finite dimensional feature maps $\phi$, such that $k (\vx, \vx') = \left \langle \phi(\vx),\phi(\vx')\right\rangle$ and the only data dependent term in KIP is $k_{t_{i}s}$ in \eqref{krr_loss}, one could consider privatizing $\phi(\vx_{t_{i}}),  \forall i \in [n]$ once at the beginning and reuse them during the training of the algorithm. However, perturbing the feature maps results in a constant signal-to-noise ratio, inversely proportional to the parameter noise $\sigma$, regardless of the number of datapoints in $\mathcal{D}_t$.  This observation implies that when the feature maps are privatized, they essentially become dominated by noise. See \suppsecref{SNR_effect} for a detailed explanation.

\section{Related Work}
The most closely related work is  \cite{DPSGD_GM}, which applies
\textit{DP-SGD} on the gradient matching  objective. As concurrent work, 
\citet{zheng2023differentially} proposes a differentially private distribution matching framework.
Our work differs from these two in the following sense. First, we use 
kernel functions as features in comparing the distilled and original data distributions. Second, we formulate our problem as kernel inducing points, our DP-SGD's privacy-accuracy trade-off is better than that of gradient matching due to privatizing a smaller dimensional quantity in our case. 

Another line of relevant work is differentially private data generation. 
While there are numerous papers to cite, we focus on a few that we compare our method against in \secref{exp}.
The majority of existing work uses DP-SGD to privatize a generator in the Generative Adversarial Networks (GANs) framework. In GANs, the generator never has direct access to training data and thus requires no privatization, as long as the discriminator is differentially private. 
Examples of this approach include 
DP-CGAN \citep{DP_CGAN},  DP-GAN \citep{DPGAN}, G-PATE \citep{g-pate}, DataLens \citep{datalens}, and GS-WGAN \cite{gs-wgan}.
A couple of recent work outside the GANs framework include 
DP-MERF \cite{dpmerf}, DP-HP \citep{DPHP}, DP-NTK \citep{dpntk_2023} and DP-MEPF \citep{harder2023pretrained} that use the approximate versions of maximum mean discrepancy (MMD) as a loss function; and DP-Sinkhorn \citep{dp_sinkhorn} that proposes using the Sinkhorn divergence in privacy settings. These data generation methods aim for more general-purpose machine learning. On the contrary, our method aims to create a privacy-preserving small dataset that matches the performance of a particular task in mind (classification). Hence, when comparing them in terms of classification performance, it is not necessarily fair for these general data generation methods. Nevertheless, at the same privacy level, we still want to show where our method stands relative to these general methods in  \secref{exp}.   

In the line of work of dataset reconstruction attacks, \citet{loo2023dataset} propose a reconstruction attack for trained neural networks that is a generalization of KIP. In addition to the task of reconstructing original training samples, they show that their attack can be used for dataset distillation when the number of samples to reconstruct is set to be significantly smaller than that of the original dataset. While the distilled samples from the reconstruction attack do not resemble any original image from the training set, there is no formal privacy guarantee about the generated images.

\section{Experiments}\label{sec:exp}

Here, we show the performance of KIP and DP-KIP over different real world datasets. In \secref{img_data} we follow previous data distillation work and focus our study on grayscale and color image datasets. In addition, we also test DP-KIP performance on imbalanced tabular datasets with numerical and categorical features in \secref{tab_data}. 
All experiments were implemented using JAX \citep{jax2018github}, except KIP e-NTK experiments were we used \texttt{autograd.grad} function implemented in PyTorch \citep{pytorch}. All the experiments were run on a single NVIDIA V100 GPU. Our code is publicly available at: \url{https://anonymous.4open.science/r/DP-KIP/}

\begin{figure}[t]
    \centering
    \includegraphics[width=0.95\linewidth]{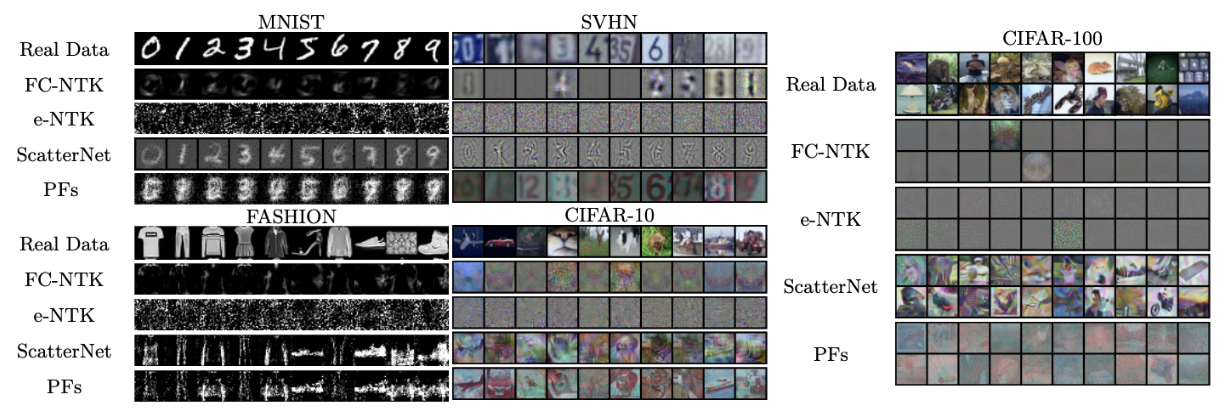}
    \caption{Generated image samples from KIP comparison models.}
    \label{fig:generated_samples}
\end{figure}


 %
  %

\subsection{Image data}\label{sec:img_data}

We start by testing KIP and DP-KIP performance on MNIST \citep{lecun2010mnist}, FashionMNIST \citep{xiao2017fashion}, SVHN \citep{netzer2011svhn}, CIFAR-10 \citep{Krizhevsky2009cifar10} and CIFAR-100 datasets for image classification. 

Each of MNIST and FashionMNIST datasets consists of 60,000 samples of $28 \times 28$ grey scale images depicting handwritten digits and items of clothing, respectively, sorted into 10 classes. 
The SVHN dataset contains 600,000 samples of $32 \times 32$ colour images of house numbers taken from Google Street View images, sorted into 10 classes. 
The CIFAR-10 dataset contains a 50,000-sample of $32 \times 32$ colour images of 10 classes of objects, such as vehicles (cars and ships) and animals (horses and birds). CIFAR100 contains $100$ classes of objects. 

In \tabref{summary_image}, we consider KIP with  infinite-width NTK for fully connected (KIP FC-NTK) and convolutional (KIP ConvNet-NTK) networks as a non-private baseline. We show as evaluation metric the averaged test accuracy using KRR as a classifier for 1, 10 and 50 distilled images per class over 5 independent runs. The general trend for different KIP variants is that creating more images per class improves the classification accuracy. The proposed e-NTK, ScatterNet and PFs features increase significantly the accuracy with respect to KIP FC-NTK for all the considered image datasets and more interestingly, KIP ScatterNet outperforms or performs similar to KIP ConvNet-NTK for most datasets. \figref{generated_samples} shows learned distilled images obtained by the different considered kernels. Both, infinite and empirical NTKs distilled images look blurrier and are visually more indistinguishable between the different classes and datasets. On the contrary, ScatterNet and PFs distilled images are visually more interpretable.

Our intuition for the interpretability of the images distilled from PFs relies on the information carried by the DCNN layer's outputs detailed in \citet{yosinski2014transferable}. DCNNs layer's outputs transfer from general information, such as color blobs in images (early layers), to specific information related to a particular class (last layers), and thus the distilled images PFs representations tend to encode all this information through KIP training. In ScatterNet features case, the intuition for the visual interpretability of the distilled samples is that the feature representations conserve the energy of the original signal (or information within the image) as described in \citet{scatternet_2015}. Therefore, the learned distilled images tend to learn more detailed patterns from the original images.

\begin{figure}[t]
    \centering
    \includegraphics[width=1.0\linewidth]{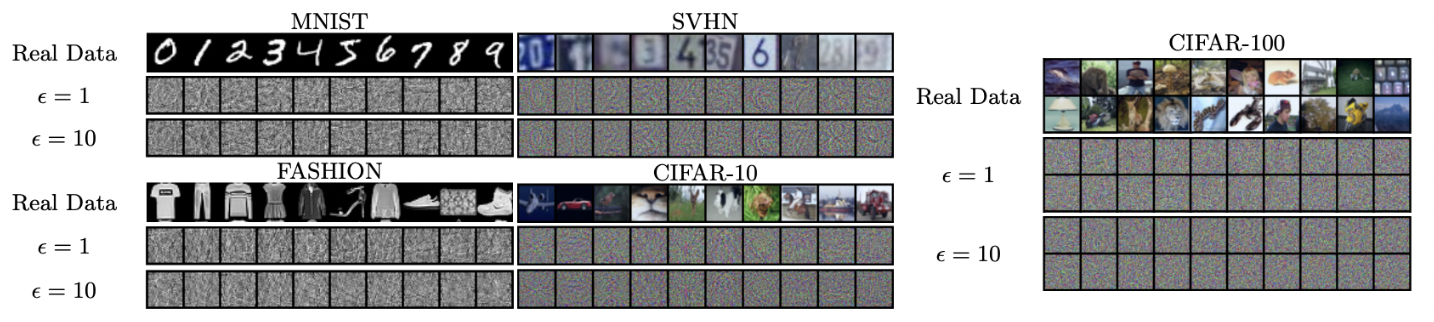}
   \caption{Generated image samples from DP-KIP-ScatterNet for different $\epsilon$ values.}
    \label{fig:scatter_dpkip_generated_samples}
\end{figure}

In the privacy realm, we evaluated DP-KIP performance in \tabref{summary_image_dp} using fully connected infinite-width NTK (DP-KIP FC-NTK) as private baseline and compared it to ScatterNet features (DP-KIP-ScatterNet). We consider $\epsilon \in \{ 1, 10\}$ and $\delta=10^{-5}$. As in the non-private setting, the general trend is that creating more images per class improves the classification accuracy, while the gradient dimension in DP-SGD increases. Furthermore, the results show that DP-KIP-ScatterNet significantly improves the performance compared to DP-KIP using FC-NTK. In \figref{scatter_dpkip_generated_samples}, we show the learned distilled images at $\epsilon=1$ and $\epsilon=10$. It is surprising that the images created at $\epsilon=10$ are not humanly discernible, while the classifiers can still achieve a good classification performance. Detailed hyperparameter settings can be found in  \suppsecref{img_data_setting}. 

\begin{table}[t!]
\caption{KRR test accuracy on Image datasets for KIP. The average over five independent runs. KIP e-NTK, KIP ScatterNet and KIP PFs outperform KIP FC-NTK. KIP ScatterNet outperforms or performs similar to KIP ConvNet-NTK. Best accuracy is marked in bold and overlapping  overlapping best accuracies are highlighted in italics.}

\centering
\scalebox{0.8}{
\begin{tabular}{c|c| c |cc |  c |  c |  c }
\toprule
 &  Imgs/ & KIP FC-NTK & \multicolumn{2}{|c |}{KIP ConvNet-NTK } & KIP e-NTK  &  KIP ScatterNet &   KIP PFs  \\ 
 & Class &  &    no aug & aug &  &  no aug & no aug \\
 \hline
\multirow{ 3}{*}{\textbf{MNIST}} & 1 & 89.3 $\pm$ 0.1 & 97.3 $\pm$ 0.1 & 96.5 $\pm$ 0.1 & 93.6 $\pm$ 0.4 & \textbf{98.2 $\pm$ 0.1} &  94.6 $\pm$ 0.6 \\
& 10 &  96.6 $\pm$ 0.1 & \textit{\textbf{99.1 $\pm$ 0.1}} & \textit{\textbf{99.1 $\pm$ 0.1}} & 96.7 $\pm$ 0.2 & \textit{99.0 $\pm$ 0.1} & 97.7 $\pm$ 0.1 \\ 
& 50 & 97.6 $\pm$ 0.1  &  \textit{99.4 $\pm$ 0.1} & \textit{\textbf{99.5 $\pm$ 0.1}} & 97.6 $\pm$ 0.1 & \textit{99.4 $\pm$ 0.1}  &  97.9 $\pm$ 0.1\\ \hline
\multirow{ 3}{*}{\textbf{FASHION}} & 1 &  80.3 $\pm$ 0.4  & 82.9 $\pm$ 0.2 & 76.7 $\pm$ 0.2  & 75.2 $\pm$ 0.3 & \textbf{87.2 $\pm$ 0.2} & 84.6 $\pm$ 0.2 \\
& 10 &  84.8 $\pm$ 0.4 &  \textbf{91.0 $\pm$ 0.1} & 88.8 $\pm$ 0.1 & 76.2 $\pm$ 0.2 & 89.5 $\pm$ 0.1 & 88.1 $\pm$ 0.4 \\ 
& 50 &  86.1 $\pm$ 0.1 & \textbf{92.4 $\pm$ 0.1} & 91.0 $\pm$ 0.1 & 79.5 $\pm$ 0.1 & 90.6 $\pm$ 0.1 & 89.3 $\pm$ 0.3   \\ \hline

\multirow{ 3}{*}{\textbf{SVHN}} & 1 & 25.4 $\pm$ 0.3 & 62.4 $\pm$ 0.2 & 64.3 $\pm$ 0.4 & 19.9 $\pm$ 0.3 & \textbf{77.4 $\pm$ 0.2} & 62.1 $\pm$ 0.2 \\
& 10 & 59.7 $\pm$ 0.5  & 79.3 $\pm$ 0.1 & 81.1 $\pm$ 0.5 & 21.1 $\pm$ 0.4 & \textbf{84.4 $\pm$ 0.1} & 81.2 $\pm$ 0.2  \\ 
& 50 & 69.7 $\pm$ 0.1 & 82.0 $\pm$ 0.1 & 84.3 $\pm$ 0.1 & 27.5 $\pm$ 0.2 & \textbf{86.4 $\pm$ 0.1} & 82.6 $\pm$ 0.2\\ \hline

\multirow{ 3}{*}{\textbf{CIFAR-10}} & 1 &  39.3 $\pm$ 1.6 & \textbf{64.7 $\pm$ 0.2} & 63.4 $\pm$ 0.1 & 32.0 $\pm$ 0.7 & 60.2 $\pm$ 0.1 & 44.7 $\pm$ 0.6 \\
& 10 &  49.1 $\pm$ 1.1 & \textbf{75.6 $\pm$ 0.2} & 75.5 $\pm$ 0.1 & 40.0 $\pm$ 0.3 & 66.2 $\pm$ 0.2 &    53.3 $\pm$ 0.5 \\ 
& 50 &  52.1 $\pm$ 0.8  &  78.2 $\pm$ 0.2 &  \textbf{80.6 $\pm$ 0.1} & 42.4 $\pm$ 0.2 & 68.5 $\pm$ 0.1 & 54.4 $\pm$ 0.3 \\ \hline

\multirow{ 3}{*}{\textbf{CIFAR-100}} & 1 & 14.5 $\pm$ 0.4 & \textbf{34.9 $\pm$ 0.1} & 33.3 $\pm$ 0.3 & 9.4 $\pm$ 0.2 & 27.4 $\pm$ 0.2 & 19.6 $\pm$ 0.3 \\
& 10 & 12.2  $\pm$ 0.2 & 47.9 $\pm$ 0.2 & \textbf{49.5 $\pm$ 0.3} & 12.5 $\pm$ 0.2 & 35.8 $\pm$ 0.1 & 23.5 $\pm$ 0.3\\ 
& 50 & 12.3 $\pm$ 0.2 & - & - &13.7 $\pm$ 0.2 & 45.7 $\pm$ 0.1 & 24.9 $\pm$ 0.4 \\ 
\bottomrule
\end{tabular}}\label{tab:summary_image}
\end{table}

\begin{table}[t!]
\caption{KRR test accuracy on Image datasets for DP-KIP FC-NTK and DP-KIP-ScatterNet at $\epsilon=1,10$ and $\delta=10^{-5}$. The average over five independent runs. DP-KIP-ScatterNet outperforms DP-KIP FC-NTK for all image datasets. 
}
\centering
\scalebox{0.8}{
\begin{tabular}{c| c| cc | cc}
\toprule
 &  Imgs/ &   \multicolumn{2}{|c}{DP-KIP FC-NTK} & \multicolumn{2}{|c}{DP-KIP-ScatterNet}    \\ 
 & Class &  $\epsilon =1$& $\epsilon=10$ &  $\epsilon =1$& $\epsilon=10$\\
 \hline
\multirow{ 3}{*}{\textbf{MNIST}} & 1  & 82.7 $\pm$ 0.1 & 85.2 $\pm$ 0.1 & \textbf{94.7 $\pm$ 0.3} & \textbf{96.1 $\pm$ 0.2} \\
& 10  & 87.5 $\pm$ 0.4  & 89.3 $\pm$ 0.3 & \textbf{96.3 $\pm$ 0.1} & \textbf{97.4 $\pm$ 0.1} \\ 
& 50  & 92.7 $\pm$ 0.2 & 93.4 $\pm$ 0.1 & \textbf{97.2 $\pm$ 0.3} & \textbf{98.1 $\pm$ 0.1} \\ \hline
\multirow{ 3}{*}{\textbf{FASHION}} & 1  &  76.9 $\pm$ 0.1    & 78.3 $\pm$ 0.1 & \textbf{77.5 $\pm$ 0.1} & \textbf{82.1 $\pm$ 0.2} \\
& 10  & 77.7 $\pm$ 0.1 & 78.7 $\pm$ 0.4 & \textbf{83.3 $\pm$ 0.2} & \textbf{86.2 $\pm$ 0.1} \\ 
& 50  & 78.8 $\pm$ 0.1 & 81.1 $\pm$ 0.1  & \textbf{84.7 $\pm$ 0.3} & \textbf{87.6 $\pm$ 0.1} \\ \hline

\multirow{ 3}{*}{\textbf{SVHN}} & 1 & 24.9 $\pm$ 0.3 &  25.2 $\pm$ 0.2  & \textbf{55.3 $\pm$ 0.6} & \textbf{68.6 $\pm$ 0.2} \\
& 10  &  40.5 $\pm$ 1.2 & 47.2 $\pm$ 0.6 & \textbf{66.4 $\pm$ 0.3} & \textbf{74.4 $\pm$ 0.2} \\ 
& 50 & 52.7 $\pm$ 0.4 & 56.6 $\pm$ 0.4 & \textbf{71.4 $\pm$ 0.4} & \textbf{76.7 $\pm$ 0.2}  \\ \hline

\multirow{ 3}{*}{\textbf{CIFAR-10}} & 1  & 36.7 $\pm$ 0.3   & 37.3 $\pm$ 0.1 & \textbf{46.6 $\pm$ 0.5} & \textbf{50.4 $\pm$ 0.4} \\
& 10 & 38.3 $\pm$ 0.3 & 39.7 $\pm$ 0.3  & \textbf{48.6 $\pm$ 0.4} & \textbf{54.4 $\pm$ 0.3} \\ 
& 50 & 40.8  $\pm$ 0.2 & 43.7 $\pm$ 0.1 &  \textbf{49.7 $\pm$ 0.5}  &  \textbf{58.7 $\pm$ 0.2}\\ \hline

\multirow{ 3}{*}{\textbf{CIFAR-100}} & 1  &  9.9  $\pm$ 0.6  & 11.1   $\pm$ 0.1 &\textbf{ 12.7 $\pm$ 0.2} & \textbf{17.5 $\pm$ 0.2} \\
& 10  &  10.1 $\pm$ 0.3 & 12.1  $\pm$ 0.4 & \textbf{14.1 $\pm$ 0.1} & \textbf{21.1 $\pm$ 0.1} \\ 
& 50  &  11.3  $\pm$ 0.3 & 13.6  $\pm$ 0.2 & \textbf{16.2 $\pm$ 0.1} & \textbf{25.2 $\pm$ 0.1} \\
\bottomrule
\end{tabular}}\label{tab:summary_image_dp}
\end{table}

\begin{table*}
    \caption{\textbf{Left:} Test accuracy of ConvNet downstream classifier trained on synthetic/distilled data with $\delta=10^{-5}$. \textbf{Right:} Test accuracy of ConvNet downstream classifier trained with  fixed $\epsilon=10$ and $\delta=10^{-5}$ and varying the number of distilled samples per class. 
    In concurrent work \citet{zheng2023differentially}, the test accuracy on the same classifier tested on $50$ MNIST distilled samples created at $\epsilon=6.12$ and $\delta=10^{-5}$
    is $97.35\%$; that on $50$  Fashion-MNIST distilled samples  created at $\epsilon=5.45$ and $\delta=10^{-5}$
    is $82.72\%$.
    }
    \label{tab:convnet_results}
    \scalebox{0.74}{
    \begin{tabular}{c |cc | cc}
\toprule
&  \multicolumn{2}{|c|}{MNIST} & \multicolumn{2}{|c}{FASHION} \\
& $\epsilon=1$  & $\epsilon=10$ & $\epsilon=1$  & $\epsilon=10$\\ \hline
DP-CGAN & 38.1 & 52.5 & 26.3 & 50.2 \\
G-PATE & 58.8 & 80.9 & 58.1 & 69.3 \\
DataLens & 71.2 & 80.7 & 64.8 & 70.6\\
GS-WGAN & 48.3 & 84.9 & 39.2 & 63.1 \\
DP-MERF & 72.7 & 85.7 & 61.2 & 72.4 \\
DP-Sinkhorn & 70.2 & 83.2 & 56.3 & 71.1 \\
\midrule
  DP-KIP FC-NTK (10 Imgs/Class) & 93.53 &  95.39 & 87.7 & 89.1\\
 DP-KIP-ScatterNet (10 Imgs/Class) &  \textbf{94.5} & \textbf{98.0}  & \textbf{88.0} & \textbf{89.3} \\
 \midrule
 \citet{DPSGD_GM} (20 Imgs/Class)  & 80.9 & 95.6 & 70.2 & 77.7 \\
 DP-KIP FC-NTK (20 Imgs/Class) & \textbf{97.78} & 97.96 & 88.3  & 90.2 \\

DP-KIP-ScatterNet (20 Imgs/Class) & 96.3 & \textbf{98.4} & \textbf{88.5}   & \textbf{90.9}  \\
\bottomrule
    \end{tabular}}
\quad
    \scalebox{0.74}{
    \begin{tabular}{c |ccc |ccc}
\toprule
&  \multicolumn{3}{|c|}{MNIST} & \multicolumn{3}{|c}{FASHION} \\

Imgs/Class & 10 & 20 & full & 10 & 20 & full \\
\hline
Real & 93.6 & 95.9 & 99.6 & 74.4 & 77.4 & 93.5 \\
DP-SGD & - & - & 96.5  &  - & - & 82.9 \\
\hline
DP-CGAN & 57.4 & 57.1 & 52.5 & 51.4 & 53.0 & 50.2 \\
G-PATE & 70.7 & 73.6 & 80.9 & 58.6 & 62.4 & 69.3 \\
DataLens & 56.5 & 66.3 & 80.7 & 61.1 & 62.7 & 70.6 \\
GS-WGAN & 83.3 & 85.5 & 84.9 & 58.7 & 59.5 & 63.1 \\

DP-MERF & 80.2 & 83.2 & 85.7 &  66.6 & 67.9 & 72.4 \\ 

DP-Sinkhorn & 74.8 & 80.5 & 83.2 & 67.4 & 68.5 & 71.1 \\
\midrule
 \citet{DPSGD_GM} & 94.9 & 95.6 & - & 75.6 & 77.7 & - \\
DP-KIP FC-NTK & 95.39 & 97.96 & - & 89.1 &  90.2 & -\\
DP-KIP-ScatterNet & \textbf{98.0} & \textbf{98.4} & - & \textbf{89.3} & \textbf{90.9} & -\\
\bottomrule
\end{tabular}}
\end{table*}

In \tabref{convnet_results}, we explore the performance of DP-KIP compared to other methods for private data generation (DP-CGAN, G-PATE, DataLens, GS-WGAN, DP-MERF, DP-Sinkhorn) and private gradient matching by \citet{DPSGD_GM}. We report the test accuracy of a ConvNet downstream classifier consisting of 3 blocks, where each block contains one Conv layer with 128 filters, Instance Normalization, ReLU activation and AvgPooling modules and a FC layer as the final output. The classifier is trained using private data and then, it is tested on real test data over 3 independent runs for each method. Here, the methods based on DP-KIP algorithm outperform existing methods at the same privacy level, achieving the best accuracy by using ScatterNet features (DP-KIP-ScatterNet). 

\begin{table}[t!]
\caption{KRR test accuracy vs. \% of corrupted pixels on CIFAR-10. Best accuracy results are obtained for DP-KIP-ScatterNet at different epsilon values and $\delta = 10^{-5} $ with 1\% and 5\% of corrupted pixels. 
}
\centering
\scalebox{0.8}{
\begin{tabular}{ c|c|c| c |c| c   }
\toprule
  Imgs/Class & Corrupted pixels (\%) & $\epsilon=0.1$& $\epsilon=1$ & $\epsilon=10$ & $\epsilon=100$ \\
 \hline
 \multirow{ 5}{*}{1} & 0 & - & 46.6 $\pm$ 0.5 & 50.4 $\pm$ 0.4 & - \\
 & 1 & \textbf{46.8 $\pm$ 0.2} & 50.4 $\pm$ 0.3 & \textbf{52.6 $\pm$ 0.4} & \textbf{54.5 $\pm$ 0.3} \\
  & 5 & 45.7 $\pm$ 0.3 & \textbf{50.9 $\pm$ 0.4} & 51.1 $\pm$ 0.5 & 53.1 $\pm$ 0.3\\
  & 10 & 43.2 $\pm$ 0.2 & 45.9 $\pm$ 0.4 & 49.9 $\pm$ 0.3 & 52.4 $\pm$ 0.6\\
  & 40  & 41.4 $\pm$ 0.3 & 45.2 $\pm$ 0.2 & 48.3 $\pm$ 0.2 & 51.0 $\pm$ 0.1\\
 & 80 & 38.4 $\pm$ 0.3 & 40.1 $\pm$ 0.3 & 45.9 $\pm$ 0.4 & 49.5 $\pm$ 0.3 \\
 \midrule
 \multirow{ 5}{*}{10} & 0 & - & 48.6 $\pm$ 0.4 & 54.4 $\pm$ 0.3 & - \\
 & 1 & 52.5 $\pm$ 0.3  & \textbf{54.9 $\pm$ 0.1}  & \textbf{55.7 $\pm$ 0.2} & \textbf{56.2 $\pm$ 0.1} \\
  & 5 & \textbf{53.9 $\pm$ 0.1} & 54.1 $\pm$ 0.1  & 54.6 $\pm$ 0.1 & 55.3 $\pm$ 0.1\\
  & 10 &  46.3 $\pm$ 0.2 &52.1 $\pm$ 0.1  &  53.6 $\pm$ 0.2 & 53.9 $\pm$ 0.1\\
  & 40  &  46.2 $\pm$ 0.3 & 47.5 $\pm$ 0.2   & 52.2 $\pm$ 0.1 & 52.7 $\pm$ 0.2\\
 & 80 & 41.6 $\pm$  0.1 &46.5 $\pm$ 0.2   & 50.2 $\pm$ 0.1 & 50.9 $\pm$ 0.1\\
 \midrule
 \multirow{ 5}{*}{50} & 0 & - & 49.76 $\pm$ 0.5 & 58.7 $\pm$ 0.2 & - \\
 & 1 & 55.2 $\pm$ 0.1 & \textbf{56.4 $\pm$ 0.1} & \textbf{59.3 $\pm$ 0.1} & \textbf{59.8 $\pm$ 0.2}\\
  & 5 & \textbf{55.4 $\pm$ 0.1} & 56.3 $\pm$ 0.1 & 59.1 $\pm$ 0.1 & 59.3 $\pm$ 0.1\\
  & 10 & 54.1 $\pm$ 0.1 & 55.2 $\pm$ 0.2 & 58.9 $\pm$ 0.1 & 59.2 $\pm$ 0.1\\
  & 40  & 47.1 $\pm$ 0.1 & 48.8 $\pm$ 0.1 & 55.7 $\pm$ 0.1 & 56.4 $\pm$ 0.1\\
 & 80 & 46.5 $\pm$ 0.2 & 47.3 $\pm$ 0.1 & 53.3 $\pm$ 0.1 & 55.2 $\pm$ 0.1\\
\bottomrule
\end{tabular}}\label{tab:cifar10_noisy_pixels}
\end{table}

\tabref{cifar10_noisy_pixels} shows the KRR test accuracy for DP-KIP-ScatterNet with pixel corruption on CIFAR-10 for generating 1, 10 and 50 distilled images per class at different $\epsilon$ values. Here, we randomly fix some pixels from the distilled images to a noisy value and perform DP-KIP-ScatterNet on the rest of the pixels. As a general trend, we observe that KRR accuracy improves over all-pixel optimization (\tabref{summary_image_dp}) at small proportions of corrupted pixels (1\%, 5\%). We also observe that 
increasing the number of distilled samples (10 and 50 images/class) enables an increase in the proportion of corrupted pixels up to 10\% compared to all-pixel optimization KRR accuracy.

\begin{table*}[ht]
\caption{Performance comparison on Tabular datasets at $\epsilon=1$ and $\delta=10^{-5}$. The average over five independent runs. For DP-KIP FC-NTK, 10 images  per class are distilled and for private synthetic methods, as many samples as the original dataset contains are generated.   
}
\centering
\scalebox{0.8}{
\begin{tabular}{l *{7}{|cc} }
\toprule
& \multicolumn{2}{ c| }{Real}  & \multicolumn{2}{ c| }{DP-CGAN}  &
\multicolumn{2}{ c| }{DP-GAN}  &
\multicolumn{2}{ c| }{{DP-MERF}} &
\multicolumn{2}{ c| }{DP-HP} & \multicolumn{2}{ c| }{DP-NTK} & \multicolumn{2}{ c }{\textbf{DP-KIP FC-NTK}}\\ 
&\multicolumn{2}{ c| }{} &   \multicolumn{2}{ c| }{($1,10^{-5}$)-DP} & \multicolumn{2}{ c| }{($1,10^{-5}$)-DP} & 
\multicolumn{2}{ c| }{($1,10^{-5}$)-DP} &
\multicolumn{2}{ c| }{($1,10^{-5}$)-DP} &
\multicolumn{2}{ c }{($1,10^{-5}$)-DP} &
\multicolumn{2}{ c }{($1,10^{-5}$)-DP}\\ 
& ROC & PRC & ROC & PRC & ROC & PRC & ROC & PRC & ROC & PRC & ROC & PRC & ROC & PRC \\
\midrule
\textbf{adult} & 0.786 & 0.683 &  0.509 & 0.444 & 0.511 & 0.445 & 0.642 & 0.524 & 0.688 & \textbf{0.632} & \textbf{0.695} & 0.557 & 0.662 & 0.365 \\
\textbf{census} & 0.776 & 0.433   & 0.655 & 0.216 & 0.529 & 0.166 & 0.685 & 0.236 & 0.699 & 0.328 & 0.71 & \textbf{0.424} & \textbf{0.766} & 0.408 \\
\textbf{cervical} & 0.959 & 0.858  & 0.519 & 0.200 & 0.485 & 0.183 & 0.531 & 0.176 & 0.616 & 0.312 & \textbf{0.631} & \textbf{0.335} & 0.622 & 0.316 \\
\textbf{credit} & 0.924 & 0.864 &  0.664 & 0.356 & 0.435 & 0.150 & 0.751 & 0.622 & 0.786 & 0.744 &  0.821 & \textbf{0.759} & \textbf{0.892} & 0.610\\
\textbf{epileptic} & 0.808 & 0.636 &  0.578 & 0.241 & 0.505 & 0.196 & 0.605 & 0.316 & 0.609 & 0.554 & 0.648 & 0.326 & \textbf{0.654} & \textbf{0.585} \\
\textbf{isolet} & 0.895 & 0.741 & 0.511 & 0.198 & 0.540 & 0.205 & 0.557 & 0.228 & 0.572 & 0.498 & 0.53 & 0.205 &  \textbf{0.615} & \textbf{0.955} \\ 

& \multicolumn{2}{ c| }{F1}
& \multicolumn{2}{ c| }{F1}
& \multicolumn{2}{ c| }{F1}
& \multicolumn{2}{ c| }{F1}
& \multicolumn{2}{ c |}{F1}
& \multicolumn{2}{ c |}{F1}
& \multicolumn{2}{ c }{F1}\\
\textbf{covtype} & \multicolumn{2}{ c| }{0.820} &
\multicolumn{2}{ c| }{0.285} &
\multicolumn{2}{ c| }{0.492} &
\multicolumn{2}{ c| }{0.467} &  \multicolumn{2}{ c |}{0.537} &
\multicolumn{2}{ c |}{0.552} &
\multicolumn{2}{ c }{\textbf{0.582}}\\
\textbf{intrusion} & \multicolumn{2}{ c| }{0.971} &  
\multicolumn{2}{ c| }{0.302} & 
\multicolumn{2}{ c| }{0.251} &
\multicolumn{2}{ c| }{\textbf{0.892}} & 
\multicolumn{2}{ c| }{0.890} &
\multicolumn{2}{ c |}{0.717} &
\multicolumn{2}{ c }{0.873} \\ 

\bottomrule
\end{tabular}}\label{tab:summary_tabular}
\end{table*}

\subsection{Tabular data}\label{sec:tab_data}

In the following we present DP-KIP results applied to eight different tabular datasets for imbalanced data. These datasets contain both numerical and categorical input features and are described in detail in \suppsecref{data_description}. To evaluate the utility of the distilled samples, we train 12 commonly used classifiers on the distilled data samples and then evaluate their performance on real data for 5 independent runs. 
In tabular data experiments, we tested our method only with infinite-width fully-connected NTKs. The motivation behind this is that by construction, ScatterNet features are designed to detect and describe local image features and thus, may not be a suitable option for tabular datasets. 

For datasets with binary labels, we use  the area under the receiver characteristics curve (ROC) and the area under the precision recall curve (PRC) as evaluation metrics, and for multi-class datasets, we use  F1 score. \tabref{summary_tabular} shows the average over the classifiers (averaged again over the 5 independent runs) trained on the synthetic privated generated samples for DP-CGAN \cite{DP_CGAN}, DP-GAN \citep{DPGAN}, DP-MERF \citep{dpmerf}, DP-HP \citep{DPHP} and DP-NTK \citep{dpntk_2023} and trained on the privately distilled samples for DP-KIP FC-NTK under the same privacy budget $\epsilon=1$ and $\delta=10^{-5}$. Details on hyperparameter settings and classifiers used in evaluation can be found in \suppsecref{tab_data_setting}.

For private synthetic methods we generate as many samples as the target dataset contains while for DP-KIP FC-NTK we set the images per class to 10. Unsurprisingly, our method outperforms the general data generation methods at the same privacy level. 


\section{Summary and Discussion}



In this paper, we propose alternative kernels to infinite-dimensional NTK in KIP algorithm. The alternative kernels reduce computational resources required to run the algorithm (down to a single GPU), while maintaining (even improving in particular cases) the quality of the distilled dataset. The computational requirement reduction has a positive impact on the carbon footprint and makes the algorithm accessible for a wider range of practitioners. Among all kernels tested, we empirically show that the kernel defined by ScatterNet features is the most suitable in image data experiments.   

As illustrated by \cite{vitaly22}, data distillation algorithms lack inherent privacy guarantees. In response to this privacy concern, we develop DP-KIP algorithm using DP-SGD on the existing KIP algorithm and implement it in JAX. 
Experimental results show that our proposed method outperforms existing data distillation methods at the same privacy level under image classification tasks with DP-KIP-ScatterNet achieving the highest performance. Additionally, we conduct experiments involving corrupted pixels on DP-KIP-ScatterNet on CIFAR10. The experiments with corrupted pixels demonstrate that fixing a small percentage of randomly selected pixels (1\%, 5\%) and applying DP-KIP-ScatterNet on the remaining pixels positively impacts the quality of the resulting distilled samples, surpassing the all-pixel optimization in terms of KRR accuracy. 

We also evaluated DP-KIP in tasks involving the classification of tabular data using the infinitely-wide fully-connected NTK. The experimental findings indicate that, despite distilling only 10 images per class, our approach typically outperforms differentially private methods for data generation, which generate samples matching the size of the target dataset. To the best of our knowledge, this is the first work that implements a data distillation algorithm to a tabular data classification problem. 

In future work, we aim to enhance our proposed method further, narrowing the accuracy gap with respect to KIP performance, especially in the context of image datasets.

\bibliography{ms}
\bibliographystyle{tmlr}

\newpage
\appendix
\begin{center}
    {\LARGE\textbf{Appendix}}
\end{center}

\section{Clipping effect of the gradients}\label{supp:clipping_effect}

In this section we empirically show the effect of clipping the gradients for DP-KIP-ScatterNet in terms of KRR accuracy. \tabref{clipping_effect} shows the accuracy for 1 Imgs/Class for MNIST at $\epsilon=1, 10$ and $\delta=10^{-5}$.

\begin{table}[h]
\caption{KRR accuracy for 1 Imgs/Class distilled dataset on MNIST at different $C$.}
\label{tab:clipping_effect}
\vskip 0.1in
\centering
\scalebox{1.0}{
\begin{tabular}{l c c c c c c}
\toprule
& \multicolumn{6}{c}{$C$ }  \\
 & $ 10^{-1} $  & $ 10^{-2} $ & $ 10^{-3} $ & $ 10^{-4} $ & $ 10^{-5} $ & $ 10^{-6} $ \\
\midrule
$\epsilon = 1$ & 81.5 $\pm$ 0.4 & \textbf{94.7 $\pm$ 0.3} & 93.7 $\pm$ 0.4 & 93.9 $\pm$ 0.4 & 93.5 $\pm$ 0.5 & 90.5 $\pm$ 0.2 \\
$\epsilon = 10$ & 80.2 $\pm$ 0.6 &  \textbf{96.1 $\pm$ 0.2} & 94.8 $\pm$ 0.2 & 94.9 $\pm$ 0.1 & 94.7 $\pm$ 0.2 & 94.2 $\pm$ 0.4 \\

\bottomrule\\
\end{tabular}

}
  \vspace{-0.5cm}
\end{table}

\section{Signal-to-Noise ratio of privatizing the feature maps}\label{supp:SNR_effect}

\begin{defn}{Signal-to-Noise Ratio (SNR) \cite{SNR_def}}
The Signal-to-Noise Ratio of a measurement can be expressed as $\frac{signal}{noise}$, where the signal and noise are expressed in the same units. An alternative way of writing this is:

\begin{equation}
    SNR = \frac{\mu}{\sigma}
\end{equation}

where $\mu$ is the signal mean or expected value and $\sigma$ is the standard deviation of the noise.
\end{defn}




Now, we want to to see what's the SNR in DP-KIP scenario for privatizing data dependent feature maps once and reuse them during training. Consider a target dataset $\mathcal{D}_{t} =  \{ (\vx_{t_{i}}, y_{t_{i}}) \}_{i=1}^{n}$ with input features $\vx_{t_{i}} \in  \mathbb{R}^{D}$ and scalar labels $y_{t_{i}}$. Let $k\colon \mathbb{R}^{D}\times \mathbb{R}^{D}$ be a positive definite kernel. By Moore--Aronszajn theorem \cite{aronszajn1950theory}, here exists
a unique reproducing kernel Hilbert space of functions on $\mathbb{R}^{D}$ for which $k$ is a reproducing kernel and thus, we can find a feature map,  $\phi\colon\mathbb{R}^{D}\to\mathcal{H}$
such that $k(\vx,\vx')=\left\langle \phi(\vx),\phi(\vx')\right\rangle _{\mathcal{H}}$ for any pair of datapoints $\vx, \vx' \in  \mathcal{X}$.

We start by deriving the sensitivity of the feature map for neighboring datasets $\mathcal{D}$ and $\mathcal{D'}$. Without loss of generality, we consider that $\mathcal{D}$ and $\mathcal{D'}$  differ in the last datapoint such that $\vx_{t_n} \neq \vx'_{t_n}$ and normalized feature maps, $\| \phi(\vx_{t_{i}}) \| = 1, \forall i \in [ n ]$. 

\begin{equation}
     \Delta_{\phi}^2 = \max_{\Dat, \Dat'} \left \| \begin{bmatrix}
 \phi(\vx_{t_1}) \\
 \phi(\vx_{t_2}) \\
 \vdots  \\
  \phi(\vx_{t_n})\\
    \end{bmatrix} - \begin{bmatrix}
 \phi(\vx_{t_1}) \\
 \phi(\vx_{t_2}) \\
 \vdots  \\
  \phi(\vx'_{t_n})\\
    \end{bmatrix} \right \|_{2} =  \max_{\Dat, \Dat'} \left \|  \phi(\vx_{t_n}) -  \phi(\vx'_{t_n}) \right \|_{2} \leq  2
\end{equation}

where the last inequality is due to the triangular inequality.



Now, we compute the SNR for the the feature map representations of  datapoints in $\mathcal{D}_{t}$:

\begin{equation}
    SNR_{\phi(\vx_{t})} = \frac{ \left \|  \begin{bmatrix}
 \phi(\vx_{t_1}) \\
 \phi(\vx_{t_2}) \\
 \vdots  \\
  \phi(\vx_{t_n})\\
\end{bmatrix}\right \|_{F}}{\sqrt{n \Delta_{\phi}^2 \sigma^2}} = \frac{\sqrt{ \| \phi(\vx_{t_1}) \|_{2}^{2} + \cdots +  \| \phi(\vx_{t_n}) \|_{2}^{2}}}{\sqrt{n \Delta_{\phi}^2 \sigma^2}} = \frac{\sqrt{n}}{2\sqrt{n}\sigma} = \frac{1}{2\sigma}
\end{equation}


Therefore, we observe that signal-to-noise ratio of the feature maps is inversevely proportional to the standard deviation parameter $\sigma$. This implies that the noise will dominate the signal-to-noise ratio unless $\sigma$ is a relatively small value which will only be the case for large $\epsilon$ values since $\sigma = \frac{\alpha \Delta_{\phi}}{\sqrt{2 \epsilon_{RDP}(\alpha)}}$ and $\epsilon (\delta) = \min_{\alpha>1}  \frac{\log(1/\delta)}{\alpha - 1} + \epsilon_{RDP}(\alpha - 1)$ for a given $\delta \ll 1/n$.


\section{Tabular datasets description}\label{supp:data_description}

\tabref{tab_data_description} contains detail information about the 8 different tabular datasets used in \secref{tab_data}. 

\begin{table}[h]
\caption{Tabular datasets. Size, number of classes and feature types descriptions.}
\label{tab:tab_data_description}
\vskip 0.1in
\centering
\scalebox{1.0}{
\begin{tabular}{l r c c}
\toprule
dataset & $\#$ samps  & $\#$ classes & $\#$ features  \\
\midrule
isolet & 4366 & 2 & 617 num \\
covtype & 406698 &  7 & 10 num, 44 cat \\
epileptic & 11500 & 2 & 178 num \\
credit & 284807 & 2 & 29 num \\
cervical & 753 & 2 & 11 num, 24 cat \\
census & 199523 & 2 & 7 num, 33 cat\\
adult & 48842 & 2 & 6 num, 8 cat\\
intrusion & 394021 & 5 & 8 cat, 6 ord, 26 num\\
\bottomrule\\
\end{tabular}

}
  \vspace{-0.5cm}
\end{table}

\section{Expermental details}

\subsection{Image data}\label{supp:img_data_setting}

Here we provide hyperparameter settings used during image data KIP experiments. \tabref{lenet_kip_settings_img}, \tabref{perceptual_kip_settings_img} and \tabref{scatter_kip_settings_img} show the number of epochs, batch size, learning rate and regularization parameter $\lambda$ used in KIP e-NTK, KIP PFs and KIP ScattterNet respectively.

\begin{table*}[h]
\caption{KIP e-NTK hyperparameter settings for Image data.}
\label{tab:lenet_kip_settings_img}
\vskip 0.1in
\centering
\begin{tabular}{l c c c c c c }
\toprule
&  Imgs/Class & epochs & batch size & learning rate & $\lambda$   \\
\midrule
\multirow{ 3}{*}{\textbf{MNIST}} & 1 & 1000 & 1000 & $5 \cdot 10^{-3}$ & $ 10^{-7} $ \\
& 10 & 1000 & 1000 & $5 \cdot 10^{-3}$ & $ 10^{-7} $  \\
& 50 & 10 & 500 & $ 10^{-2} $ & $ 10$   \\
\midrule
\multirow{ 3}{*}{\textbf{FASHION}} & 1 & 100 & 100 & $10^{-3}$ & $10^{-2}$  \\
& 10 & 100 & 100 &  $10^{-3}$ & $ 10^{-1} $ \\
& 50 & 10 & 100 &  $ 10^{-2} $ & $ 10^{-1} $ \\
\midrule
\multirow{ 3}{*}{\textbf{SVHN}} & 1 & 100 & 2000 & $10^{-2} $ & $10^{-3}$  \\
& 10 & 100 & 2000 & $10^{-2}$ & $10^{-4}$ \\
& 50 & 10 & 1000 & $10^{-4}$ & $10^{-3}$\\
\midrule
\multirow{ 3}{*}{\textbf{CIFAR-10}} & 1 & 100 & 200 & $10^{-2}$ &  $10^{-3}$   \\
& 10  & 100 &  1000 & $10^{-2}$ & $10^{-4}$  \\
& 50 & 100 & 2000 &  $10^{-2}$  &  $10^{-2}$ \\
\midrule
\multirow{2}{*}{\textbf{CIFAR-100}}& 1 & 100 &  1000 & $10^{-1}$ & $10^{-3}$ \\ 
& 10 & 1000 & 50 & $10^{-1}$ & $10^{-3}$  \\
& 50 & 1000 & 50 & $10^{-1}$ & $10^{-2}$  \\
\bottomrule
\end{tabular}
\end{table*}

\begin{table*}[ht]
\caption{KIP PFs hyperparameter settings for Image data.}
\label{tab:perceptual_kip_settings_img}
\vskip 0.1in
\centering
\begin{tabular}{l c c c c c c }
\toprule
&  Imgs/Class & epochs & batch size & learning rate & $\lambda$   \\
\midrule
\multirow{ 3}{*}{\textbf{MNIST}} & 1 & 100 & 1000 & $10^{-2}$ & $10^{-6}$  \\
& 10 & 100 & 5000 & $ 10^{-2} $ & $ 10^{-5} $  \\
& 50 & 100 & 2000 & $ 10^{-2} $ & $ 10^{-6} $   \\
\midrule
\multirow{ 3}{*}{\textbf{FASHION}} & 1 & 1000 & 2000 & $10^{-2}$ & $ 10^{-5} $  \\
& 10 & 1000 & 1000 &  $10^{-2}$ & $ 10^{-5} $ \\
& 50 & 1000 & 5000 &  $ 10^{-3} $ & $ 10^{-5} $ \\
\midrule
\multirow{ 3}{*}{\textbf{SVHN}} & 1 & 100 & 2000 & $10^{-2} $ & $10^{-3}$  \\
& 10 & 100 & 2000 & $10^{-2}$ & $10^{-3}$ \\
& 50 & 1000 & 100 & $10^{-4}$ & $10^{-2}$\\
\midrule
\multirow{ 3}{*}{\textbf{CIFAR-10}} & 1 & 1000 & 2000 & $10^{-2}$ &  $10^{-4}$   \\
& 10  & 1000 &  5000 & $10^{-3}$ & $10^{-3}$  \\
& 50 & 1000 & 2000 &  $10^{-3}$  &  $10^{-2}$ \\
\midrule
\multirow{2}{*}{\textbf{CIFAR-100}}& 1 & 1000 &  1000 & $10^{-2}$ & $10^{-3}$ \\ 
& 10 & 1000 & 200 & $10^{-2}$ & $10^{-1}$  \\
& 50 & 100 & 200 & $10^{-2}$ & $10^{-3}$  \\
\bottomrule
\end{tabular}
\end{table*}

\begin{table*}[ht]
\caption{KIP ScatterNet hyperparameter settings for Image data.}
\label{tab:scatter_kip_settings_img}
\vskip 0.1in
\centering
\begin{tabular}{l c c c c c c }
\toprule
&  Imgs/Class & epochs & batch size & learning rate & $\lambda$   \\
\midrule
\multirow{ 3}{*}{\textbf{MNIST}} & 1 & 1000 & 200 & $10^{-4}$ & $1$  \\
& 10 & 2000 & 100 & $ 10^{-4} $ & $ 10^{-5} $  \\
& 50 & 2000 & 5000 & $ 10^{-4} $ & $ 10^{-5} $   \\
\midrule
\multirow{ 3}{*}{\textbf{FASHION}} & 1 & 1000 & 1000 & $10^{-3}$ & $ 1 $  \\
& 10 & 2000 & 2000 &  $10^{-3}$ & $ 10^{-6} $ \\
& 50 & 2000 & 2000 &  $ 5 \cdot 10^{-3} $ & $ 10^{-2} $ \\
\midrule
\multirow{ 3}{*}{\textbf{SVHN}} & 1 & 1000 & 200 & $10^{-4} $ & $1$  \\
& 10 & 1000 & 1000 & $10^{-3}$ & $10^{-4}$ \\
& 50 & 8000 & 5000 & $ 5 \cdot 10^{-3}$ & $10^{-3}$\\
\midrule
\multirow{ 3}{*}{\textbf{CIFAR-10}} & 1 & 1000 & 2000 & $10^{-3}$ &  $1$   \\
& 10  & 1000 &  2000 & $10^{-3}$ & $10^{-6}$  \\
& 50 & 1000 & 5000 &  $10^{-3}$  &  $10^{-3}$ \\
\midrule
\multirow{2}{*}{\textbf{CIFAR-100}}& 1 & 10000 &  2000 & $10^{-2}$ & $10^{-3}$ \\ 
& 10 & 1000 & 5000 & $10^{-2}$ & $10^{-3}$  \\
& 50 & 2000 & 1000 & $10^{-2}$ & $10^{-1}$  \\
\bottomrule
\end{tabular}
\end{table*}

Here we provide the details of the DP-KIP training procedure we used on the image data experiments. \tabref{dpkip_settings_img_eps1} and \tabref{dpkip_settings_img_eps10} show the number of epochs, batch size, learning rate, clipping norm $C$ and regularization parameter $\lambda$ used during training for each image classification dataset at the corresponding images per class distilled for DP-KIP FC-NTK. \tabref{scatter_dpkip_settings_img_eps1} and \tabref{scatter_dpkip_settings_img_eps10} show the different hyperparameter settings used in DP-KIP-ScatterNet experiments.

\begin{table*}[ht]
\caption{DP-KIP FC-NTK hyperparameter settings for Image data for $\epsilon=1$ and $\delta=10^{-5}$.}
\label{tab:dpkip_settings_img_eps1}
\vskip 0.1in
\centering
\begin{tabular}{l c c c c c c c }
\toprule
&  Imgs/Class & epochs & batch size & learning rate & $C$ & $\lambda$   \\
\midrule
\multirow{ 3}{*}{\textbf{MNIST}} & 1 & 10 & 500 & $5\cdot10^{-3}$ & $10^{-2}$ &$10^{-6}$  \\
& 10 & 10 & 500 & $ 10^{-2} $ & $ 10^{-6} $ & $ 10^{-6} $  \\
& 50 & 10 & 500 & $ 10^{-2} $ & $ 10^{-6} $ & $ 10^{-7} $  \\
\midrule
\multirow{ 3}{*}{\textbf{FASHION}} & 1 & 10 & 500 & $5\cdot10^{-2}$ & $ 10^{-6} $ & $ 10^{-6} $  \\
& 10 & 10 & 500 &  0.1 & $ 10^{-6} $ & $ 10^{-5} $\\
& 50 & 10 & 200 &  $ 10^{-2} $ & $ 10^{-2} $ & $ 10^{-6} $\\
\midrule
\multirow{ 3}{*}{\textbf{SVHN}} & 1 & 10 & 50 & $10^{-1} $ & $10^{-6}$ & $ 10^{-6} $ &  \\
& 10 & 10 & 500 & $5\cdot10^{-2}$ & $10^{-6}$ & $10^{-6}$\\
& 50 & 10 & 500 & $10^{-2}$ & $10^{-5}$ & $10^{-2}$\\
\midrule
\multirow{ 3}{*}{\textbf{CIFAR-10}} & 1 & 20 & 200 & $10^{-2}$ &  $10^{-4}$ &  $10^{-5}$  \\
& 10  & 10 &  500 & $5\cdot10^{-2}$ & $10^{-5}$ & $10^{-6}$ \\
& 50 & 10 & 500 &  $5\cdot10^{-2}$  &  $10^{-3}$ &  $10^{-6}$\\
\midrule
\multirow{2}{*}{\textbf{CIFAR-100}}& 1 & 10 &  200 & $10^{-2}$ & $10^{-5}$ & $10^{-7}$ \\ 
& 10 & 10 & 100 & $10^{-2}$ & $10^{-4}$ & $10^{-7}$ \\
& 50 & 10 & 50 & $10^{-2}$ & $10^{-3}$ & $10^{-7}$ \\
\bottomrule
\end{tabular}
\end{table*}

\begin{table*}[ht]
\caption{DP-KIP FC-NTK hyperparameter settings for Image data for $\epsilon=10$ and $\delta=10^{-5}$.}
\label{tab:dpkip_settings_img_eps10}
\vskip 0.1in
\centering
\begin{tabular}{l c c c c c c c }
\toprule
&  Imgs/Class & epochs & batch size & learning rate & $C$ & $\lambda$   \\
\midrule
\multirow{ 3}{*}{\textbf{MNIST}} & 1 & 10 & 500 & $5\cdot10^{-3}$ & $10^{-5}$ &$10^{-6}$  \\
& 10 & 10 & 500 & $5\cdot10^{-3}$ & $10^{-2}$ & $10^{-6}$ \\
& 50 & 10 & 500 & $5\cdot10^{-3}$ & $10^{-5}$ & $10^{-6}$   \\
\midrule
\multirow{ 3}{*}{\textbf{FASHION}} & 1 & 10 & 500 & $10^{-2}$ & $10^{-3}$ & $10^{-6}$  \\
& 10 & 10 & 500 & $10^{-2}$ & $10^{-2}$ & $10^{-5}$  \\
& 50 &  10 & 500 & $10^{-2}$ & $10^{-2}$ & $10^{-7}$  \\
\midrule
\multirow{ 3}{*}{\textbf{SVHN}} & 1 & 10 & 50 & $5\cdot10^{-2}$ &  $10^{-6}$ &  $10^{-6}$ \\
& 10 & 10 & 500 & $10^{-1}$ &  $10^{-6}$ &  $10^{-6}$  \\
& 50  & 10 & 500 & $10^{-2}$ &  $10^{-2}$ &  $10^{-7}$ \\
\midrule
\multirow{ 3}{*}{\textbf{CIFAR-10}} & 1 & 10 &  500 & $10^{-1}$ &  $10^{-6}$ &  $10^{-6}$  \\
& 10 & 10 & 500 & $5\cdot10^{-2}$ & $10^{-5}$ &  $10^{-6}$  \\
& 50 & 10 & 500 & $5\cdot10^{-2}$ & $10^{-5}$ &  $10^{-6}$ \\
\midrule
\multirow{2}{*}{\textbf{CIFAR-100}}& 1 & 10 & 100 & $10^{-2} $ & $10^{-5} $& $10^{-6} $ \\ 
& 10 & 10 & 100 & $10^{-2} $ & $10^{-4} $& $10^{-6} $ \\
& 50 & 10 & 100 & $10^{-2} $ & $10^{-3} $& $10^{-6} $ \\
\bottomrule
\end{tabular}
\end{table*}

\begin{table*}[ht]
\caption{DP-KIP-ScatterNet hyperparameter settings for Image data for $\epsilon=1$ and $\delta=10^{-5}$.}
\label{tab:scatter_dpkip_settings_img_eps1}
\vskip 0.1in
\centering
\begin{tabular}{l c c c c c c c }
\toprule
&  Imgs/Class & epochs & batch size & learning rate & $C$ & $\lambda$   \\
\midrule
\multirow{ 3}{*}{\textbf{MNIST}} & 1 & 20 & 5000 & $10^{-1}$ & $10^{-2}$ &$10^{-4}$  \\
& 10 & 30 & 500 & $ 10^{-1} $ & $ 10^{-7} $ & $ 10^{-3} $  \\
& 50 & 30 & 200 & $ 10^{-1} $ & $ 10^{-2} $ & $ 10^{-2} $  \\
\midrule
\multirow{ 3}{*}{\textbf{FASHION}} & 1 & 50 & 2000 & $10^{-1}$ & $ 10^{-4} $ & $ 10^{-7} $  \\
& 10 & 40 & 1000 &  $10^{-2}$ & $ 10^{-4} $ & $ 10^{-3} $\\
& 50 & 20 & 200 &  $ 10^{-2} $ & $ 10^{-5} $ & $ 10^{-2} $\\
\midrule
\multirow{ 3}{*}{\textbf{SVHN}} & 1 & 50 & 2000 & $10^{-1} $ & $10^{-4}$ & $ 10^{-7} $ &  \\
& 10 & 40 & 1000 & $10^{-2}$ & $10^{-4}$ & $10^{-3}$\\
& 50 & 20 & 200 & $10^{-2}$ & $10^{-5}$ & $10^{-2}$\\
\midrule
\multirow{ 3}{*}{\textbf{CIFAR-10}} & 1 & 40 & 2000 & $10^{-1}$ &  $10^{-6}$ &  $10^{-6}$  \\
& 10  & 20 &  500 & $10^{-1}$ & $10^{-6}$ & $10^{-3}$ \\
& 50 & 30 & 100 &  $10^{-2}$  &  $10^{-2}$ &  $10^{-1}$\\
\midrule
\multirow{2}{*}{\textbf{CIFAR-100}}& 1 & 30 &  200 & $10^{-4}$ & $10^{-6}$ & $10^{-2}$ \\ 
& 10 & 10 & 50 & $10^{-4}$ & $10^{-6}$ & $10^{-1}$ \\
& 50 & 10 & 50 & $10^{-4}$ & $10^{-6}$ & $10^{-2}$ \\
\bottomrule
\end{tabular}
\end{table*}

\begin{table*}[ht]
\caption{DP-KIP-ScatterNet hyperparameter settings for Image data at $\epsilon=10$ and $\delta=10^{-5}$.}
\label{tab:scatter_dpkip_settings_img_eps10}
\vskip 0.1in
\centering
\begin{tabular}{l c c c c c c c }
\toprule
&  Imgs/Class & epochs & batch size & learning rate & $C$ & $\lambda$   \\
\midrule
\multirow{ 3}{*}{\textbf{MNIST}} & 1 & 30 & 500 & $10^{-1}$ & $10^{-2}$ & $10^{-6}$  \\
& 10 & 50 & 2000 & $ 10^{-1} $ & $ 10^{-6} $ & $ 10^{-4} $  \\
& 50 & 30 & 200 & $ 10^{-1} $ & $ 10^{-7} $ & $ 10^{-1} $  \\
\midrule
\multirow{ 3}{*}{\textbf{FASHION}} & 1 & 50 & 1000 & $10^{-1}$ & $ 10^{-2} $ & $ 10^{-4} $  \\
& 10 & 50 & 2000 &  $10^{-2}$ & $ 10^{-1} $ & $ 10^{-3} $\\
& 50 & 50 & 200 &  $ 10^{-2} $ & $ 10^{-2} $ & $ 10^{-3} $\\
\midrule
\multirow{ 3}{*}{\textbf{SVHN}} & 1 & 50 & 2000 & $10^{-2}$ &  $10^{-2}$ &  $10^{-6}$  \\
& 10  & 50 &  500 & $10^{-2}$ & $10^{-5}$ & $10^{-3}$ \\
& 50 & 50 & 100 &  $10^{-2}$  &  $10^{-2}$ &  $10^{-3}$\\
\midrule
\multirow{ 3}{*}{\textbf{CIFAR-10}} & 1 & 50 & 1000 & $10^{-2}$ &  $10^{-2}$ &  $10^{-5}$  \\
& 10  & 50 &  500 & $10^{-2}$ & $10^{-4}$ & $10^{-4}$ \\
& 50 & 50 & 100 &  $10^{-2}$  &  $10^{-5}$ &  $10^{-4}$\\
\midrule
\multirow{2}{*}{\textbf{CIFAR-100}}& 1 & 50 &  1000 & $10^{-1}$ & $10^{-6}$ & $10^{-5}$ \\ 
& 10 & 50 & 50 & $10^{-1}$ & $10^{-6}$ & $10^{-2}$ \\
& 50 & 50 & 50 & $10^{-1}$ & $10^{-6}$ & $10^{-2}$ \\
\bottomrule
\end{tabular}
\end{table*}

\subsection{Tabular data}\label{supp:tab_data_setting}

\tabref{tab_data_hyper} contains DP-KIP FC-NTK hyperparameter settings used in tabular data experiments. \tabref{tab_downstream_hyperparam} describes the hyperparameter setting for training the downstream classifiers used in tabular data experiments.

\begin{table}[ht]
\caption{DP-KIP FC-NTK hyperparameter settings for tabular data.}
\label{tab:tab_data_hyper}
\vskip 0.1in
\centering
\scalebox{1.0}{
\begin{tabular}{l c c c c c c}
\toprule
dataset & epochs  & batch size (\%) & learning rate & $C$ & $\lambda$  & subsampling rate  \\
\midrule
isolet & 10 & 0.8 & $10^{-3}$ & $10^{-3}$ & $10^{-6}$ & $10^{-2}$ \\
covtype & 10 & 0.02 & $10^{-1}$ & $10^{-2}$ & $10^{-7}$ & $7 \cdot 10^{-2}$\\
epileptic & 10 & 0.07 & $10^{-2}$ & $10^{-3}$ & $10^{-6}$ & $3 \cdot 10^{-1}$ \\
credit & 10 & 0.07 & $10^{-2}$ & $10^{-1}$ & $10^{-6}$ & $10^{-2}$  \\
cervical & 10 & 0.6 & $10^{-4}$ & $10^{-3}$ & $10^{-6}$ & $2 \cdot 10^{-2}$ \\
census  & 10 & 0.4 & $5 \cdot 10^{-1}$ & $10^{-1}$ & $10^{-6}$ & $4 \cdot 10^{-2}$\\
adult & 10 & 0.8 & $10^{-2}$ & $10^{-1}$ & $10^{-6}$ & $7 \cdot 10^{-2}$\\
intrusion & 10 & 0.05 & $10^{-4}$ & $10^{-4}$ & $10^{-6}$ & $10^{-2}$\\
\bottomrule\\
\end{tabular}

}
  \vspace{-0.5cm}
\end{table}

\begin{table}[!ht]
\caption{Hyperparameter settings for downstream classifiers used in tabular data experiments. Models are taken from scikit-learn 0.24.2 and xgboost 0.90 python packages and hyperparameters have been set to achieve reasonable accuracy while limiting execution time. Parameters not listed are kept as default values.}
\label{tab:tab_downstream_hyperparam}
\vskip 0.1in
\centering
\scalebox{0.85}{
\begin{tabular}{l|l}
\toprule
Model & Parameters \\
\midrule
Logistic Regression & solver: lbfgs, max\_iter: 5000, multi\_class: auto
\\ 
Gaussian Naive Bayes & - \\
Bernoulli Naive Bayes & binarize: 0.5\\
LinearSVC & max\_iter: 10000, tol: 1e-8, loss: hinge\\
Decision Tree & class\_weight: balanced \\
LDA & solver: eigen, n\_components: 9, tol: 1e-8, shrinkage: 0.5\\
Adaboost & n\_estimators: 1000, learning\_rate: 0.7, algorithm: SAMME.R\\
Bagging & max\_samples: 0.1, n\_estimators: 20\\
Random Forest & n\_estimators: 100, class\_weight: balanced\\
Gradient Boosting & subsample: 0.1, n\_estimators: 50\\
MLP & - \\
XGB & colsample\_bytree: 0.1, objective: multi:softprob, n\_estimators: 50\\
\bottomrule
\end{tabular}
}
  \vspace{-0.5cm}
\end{table}


\end{document}